\documentclass[conference,letter,10pt,final]{llncs}
\usepackage[utf8]{inputenc}
\usepackage[T1]{fontenc}
\usepackage{graphicx}
\usepackage{longtable}
\usepackage{wrapfig}
\usepackage{rotating}
\usepackage[normalem]{ulem}
\usepackage{amsmath}
\usepackage{amssymb}
\usepackage{capt-of}
\usepackage{hyperref}
\usepackage[utf8]{inputenc}
\usepackage[T1]{fontenc}
\date{}
\title{}
\hypersetup{
 pdfauthor={nil},
 pdftitle={},
 pdfkeywords={},
 pdfsubject={},
 pdfcreator={Emacs 28.1 (Org mode 9.6)}, 
 pdflang={British}}
\begin{document}

\title{ $\mu$Boost: An Effective Method for Solving Indic Multilingual Text Classification Problem}

\author{Manish Pathak\inst{1} \and Aditya Jain\inst{2}}
\institute{
MiQ Digital, Bengaluru, India \\
\email{manishpathak@miqdigital.com}
\and
MiQ Digital, Bengaluru, India \\
\email{aditya@miqdigital.com}}


\maketitle

\begin{abstract}
Text Classification is an integral part of many Natural Language Processing tasks such as sarcasm detection, sentiment analysis and many more such applications. Many e-commerce websites, social-media/entertainment platforms use such models to enhance user-experience to generate traffic and thus, revenue on their platforms. In this paper, we are presenting our solution to Multilingual Abusive Comment Identification Problem on Moj, an Indian video-sharing social networking
service, powered by ShareChat. The problem dealt with detecting abusive
comments, in 13 regional Indic languages such as Hindi, Telugu, Kannada etc., on the videos on Moj platform. Our solution utilizes the novel \(\mu\)Boost, an ensemble of CatBoost classifier models and Multilingual Representations for Indian Languages (MURIL) model, to produce SOTA performance on Indic text classification tasks. We were able to achieve a mean F1-score of 89.286 on the test data, an improvement over baseline MURIL model with a F1-score of 87.48.

\end{abstract}

\section{Introduction}
\label{sec:org4584faa}

With internet becoming more and more accessible to users there is an exponential
growth in number of websites and platforms across all the sectors and markets.
On one hand we have some juggernauts in e-commerce space selling useful products
to customers and on the other hand we have social media/entertainment platforms
influencing the way people live and even behave. While there’s a lot of work
going on to increase the engagement of users on these platforms and thus
directly affecting the revenue growth, there’s relatively less work going on to
regulate the content shared on such platforms. In recent times, people and
organizations have become more cognizant of shared responsibility of tackling
racial/ethnic discrimination and detecting offensive and abusive language
written on online platforms.  This competition, hosted on Kaggle, was structured
around the same thought. The goal of the challenge was to find abusive comments
using natural language processing on popular Indian short-video sharing app Moj,
owned by parent company ShareChat. The training data was gold standard human
annotated. The unique thing about the data provided was that the comments
were multilingual in 13 different regional Indic languages such as Kannada,
Tamil, Telugu, Hindi, Gujarati, Bhojpuri, Marathi, Malayalam, Bengali, Odia,
Haryanvi, Rajasthani, and Assamese. Apart from this, there were 5 other
independent features such as number of likes, number of reports etc., that
provided meta-data about the comments.

\section{Data}
\label{sec:org3e5802c}

The dataset consisted of the following columns:

\begin{itemize}
\item language              : Language of the post on which this comment was made.
\item post\_index            : Unique post identifier having this comment.
\item commentText           : Comment as a string
\item report\_count\_comment  : Number of times comment has been reported.
\item report\_count\_post     : Total number of reports on all comments of this post.
\item like\_count\_comment    : Number of likes on the comment
\item like\_count\_post       : Total number of likes on all comments of this post
\item label                 : 0-Non-abusive; 1-abusive
\end{itemize}
\subsection{Exploratory Data Analysis}
\label{sec:org4274978}
The training dataset had 665042 rows and 7 independent features (X) and 1 binary
target feature \emph{`label'} indicating whether the comment is abusive or not. 52.98\%
of comments were labeled as non-abusive and 47.02\% of comments were labeled as abusive
indicating a balanced class dataset. The average number of words in the comments, tokenized on
space, were 12.7. Around 46\% of the comments were in Hindi/Hinglish, 14.5\% in
Telugu, 10.8\% in Marathi, 6.1\% in Malayalam, 3.4\% in Bengali, 2\% in Kannada,
1.6\% in Odia, 1.3\% in Gujarati, 1.3\% in Haryanvi, 0.8\% in Bhojpuri, 0.6\% in
Rajasthani and 0.4\% in Assamese. There were 391117 unique post indexes. The
distribution of other features are presented in table  \ref{eda}.

\begin{table*}[htbp]
\caption{EDA Summary  \label{eda}}
\centering
\begin{tabular}{lrrr}
Column Name & Mean & Standard deviation & Maximum\\
\hline
report\_count\_comment  & 0.0052 & 0.08 & 8\\
report\_count\_post  & 0.2171 & 13.84 & 3359\\
like\_count\_comment  & 0.6865 & 7 & 2344\\
like\_count\_post  & 255.9552 & 2548.12 & 104159\\
\end{tabular}
\end{table*}

\section{Embedding based approaches}
\label{sec.proposal}
We tried multiple approaches before comming up with \(\mu\)-boost. In this section we
will briefly describe various approaches that we have taken and their results.

We have split the data as 99\% train and 1\% dev set. This split is used across
all experiments.
\subsection{BiLSTM and BiGRU}
\label{sec:org9a95621}
Recurrent neural network models have been known to perform very well on NLP tasks\cite{schuster1997}. Based on this
understanding we built our baseline model using 2 layers of BiLSTM\cite{schuster1997,hochreiter1997}. The first
BiLSTM layer was provided with embeddings learnt by Keras' Embedding layer
having a dimension of 64x120. For this experiment we relied on Keras' inbuilt tokenizer.
One thing to note here is that we only resorted to
the text column in the input data. We did not use any other column from the
provided dataset.  The number of LSTM nodes used were 64, and 32 in first and
second layers respectively. The baseline F1 score obtained by this model on the
training set was 87.515 and on the test set was 86.432. This model also showed
signs of overfitting in very early stages of training. We also experimented with
replacing LSTM units with GRU units in the BiLSTM layer. The F1 score  of BiGRU
model on the training set was 88.511 and on the test set was 86.560.
We calculated these F1 scores at the probability threshold of 0.5.
Both these models were trained on Google Colab's 12 GB GPU-RAM instance and took approximately 1 hour to
train. The next logical step was to use pretrained models or word embeddings.
\subsection{MuRIL}
\label{sec:org055dd74}

MuRIL stands for Multilingual Representations for Indian Languages. This
approach is currently the SOTA for Indian Language models. It out-performs mBert
on Indian Languages. MuRIL uses transformers \cite{khanuja2021} as building blocks of the
model.  Transformers incorporate self-attention in the encoder layers, and both
self-attention and encoder attention in the decoder layers. This gives
transformers an edge over recurrent models like LSTMs which receive input one
word/step at a time and have difficulty in learning very long term dependencies
which occur in natural language.

The differentiating factor of this model with respect to BERT is that it is
trained only on data of Indic Languages, and is trained on both translated and
transliterated datasets.

\begin{figure}[htbp]
\centering
\includegraphics[width=.9\linewidth]{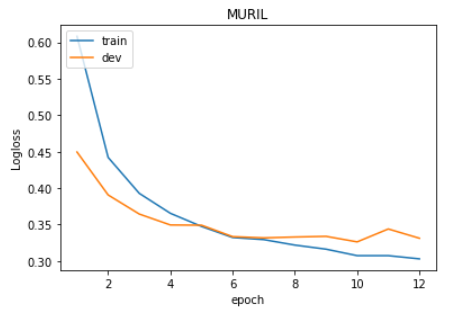}
\caption{MuRIL Training and Dev Loss}
\end{figure}

MuRIL covers 17 languages, however, some of the languages required by the
competition are not covered by MuRIL for which we will discuss our solution in the
next section.  A pretrained model for MuRIL is generously made available by the
MuRIL authors on huggingface, which we have used in this experiment.

We relied on MuRIL's tokenizer for processing raw text. This model was trained for 12 \emph{epochs} with \emph{max\_len} of 64 and batch size of 128.
We also used a dropout layer on the output of transformer layer with dropout ratio of 0.2 to regularize the model.
The optimizer used was \emph{ADAM} with a learning rate of 1e-5.

Our approach to using MuRIL is inspired by the Kaggle notebook \cite{zotero-221} that had a
test F1 score of 87.48 with a corresponding training F1 score of 88.353 at probability threshold of 0.5.  It
took us approximately 4.5 hours to train this model on default kaggle TPU
instance.

\section{Bag-of-words based approaches}
\label{sec:orgd8cc2e1}

In order to improve the above MURIL model in a world where powerful deep
learning algorithms are ruling the arena, we chose to train a simple Bag of
Words model \cite{zhang2010}.
\subsection{Choice of classifier}
\label{sec:orgf5cc2f1}
\begin{table*}[htbp]
\caption{Comparison of various models \label{perf}}
\centering
\begin{tabular}{lrrrrrr}
 & BiLSTM & BiGRU & Single CatBoost & Ensemble CatBoost & MuRIL & \(\mu\)Boost\\
\hline
Train F1 & 87.515 & 88.511 & 92.171 & 92.961 & 88.35 & 92.831\\
Test F1 & 86.432 & 86.560 & 87.153 & 87.409 & 87.48 & 89.282\\
\end{tabular}
\end{table*}
As discussed in previous sections, predicting a comment as abusive or
non-abusive is a typical text classification task. In order to build this model,
we had to decide which classifier to choose. We chose CatBoost as our
classifier, which is based on the principles of Gradient-Boosted tree framework,
to predict the probabilities of a comment being abusive. The dataset was
structured in   nature with categorical columns such as \emph{post\_index} with a very
high cardinality.

\begin{table}[htbp]
\caption{Ordered Target Encoding \label{ote}}
\centering
\begin{tabular}{lr}
language & label\\
\hline
Hindi & 1\\
Hindi & 0\\
Marathi & 0\\
\(Hindi_i\) & 1\\
\end{tabular}
\end{table}

The most common approach to handle categorical columns is to
do one-hot encoding \cite{cohen2014a,chapelle2015,micci-barreca2001a} but that would have led to creation of high
dimensional feature vectors which can worsen model performance and increase
training time. Other methods such as target encoding \cite{micci-barreca2001a} are also sensitive to
target leakage. Classifiers such as XGBoost has shown great performance on
variety of structured datasets, but they consume more memory as compared to
CatBoost\cite{hancock2020}.

CatBoost models can also be trained very easily on GPUs by passing an argument
\emph{task\_type} as GPU. Another reason for choosing CatBoost as the classifier was
its ability to handle categorical variables without the risk of target
leakage\cite{prokhorenkova2019}. CatBoost performs Ordered Target encoding to encode categorical
variables. To encode a particular category via ordered target encoding CatBoost
utilizes label information of the respective categories’ instances that occurred
prior to current instance. For example, in the table \(language\) feature
category \(Hindi_i\), where \(i\) denotes the \(i^{th}\)  instance of that category, will
get encoded by the formula \ref{eq.ode}. It also assigns a prior value to avoid
undefined values in encoding.

\begin{equation}
\label{eq.ode}
Hindi_i \rightarrow  \frac{1+0+a*Prior}{2+a} \end{equation}

\subsection{Text features in CatBoost}
\label{sec:org7399fc6}
Apart from handling categorical features, CatBoost can also handle text features
internally quite well via the \(text\_features\) class. It provides variety of text
preprocessing steps on the fly. It can tokenize and lowercase the text by
splitting each sentence into words or letters. Then it creates a dictionary that
collects all values of text feature and numbers the minimum unit of text
sequence representation called a token. Apart from this it also provides control
of n-gram tokenization as combining tokens can be useful to perceive contiguous
text. It is also possible to filter rare occurring words and specify the maximum
dictionary size\cite{zotero-225}. The text features are represented as
numerical features by Bag of Words (BoW) representation where a Boolean flag
reflects whether the text contains the token or not. This has the problem of
sparse representation as compared to other embedding vector representations that
are generally used in deep learning models, but it can be controlled and tuned
by varying the \emph{max\_dictionary\_size} and \emph{top\_tokens\_count} hyperparameters.
These computed sparse numerical features are then fed into regular CatBoost
training algorithm.  To begin with, we trained a simple CatBoost classifier on
Google Colab’s 12 GB RAM Nvidia GPU. We kept 99\% of the data in train set and
rest 1\% in development set. Categorical features that were fed into CatBoost
were the columns \emph{language} and \emph{post\_index}. The feature \emph{commentText} was
passed as the text feature. The \emph{max\_dictionary\_size} was set to 800000 and
\emph{top\_tokens\_count} to only 16000. We trained the model for 15000 iterations with
\emph{F1} score as the evaluation metric. CatBoost also incorporates an overfitting
detector, \emph{od\_wait}, and can stop training if score doesn’t improve for
specified number of iterations. We kept the \emph{od\_wait} hyperparameter as 2000
iterations. The \emph{learning\_rate} was set to 0.35 and \emph{depth} as 12. The model
gave train F1 score of 92.171 and test F1 score of 87.154 at 0.5 probability
threshold.
\subsection{Ensembling CatBoost Models}
\label{sec:orga42e370}
Continuing our experiments, we next explored the effect of ensembling of
CatBoost models on the performance metrics. Here, we trained 3 CatBoost models
in a loop with different seed parameters but same configurations in order to
predict the probabilities of abusive comments. The time taken to train this
ensemble model with overfitting detector was approximately 1.45 hours. The
probabilities were then averaged across the three trained models to get final
prediction probabilities. The train F1 score increased to 92.961 and test F1
score to 87.409, with a probability threshold of 0.5.

\begin{figure}[htbp]
\centering
\includegraphics[width=.9\linewidth]{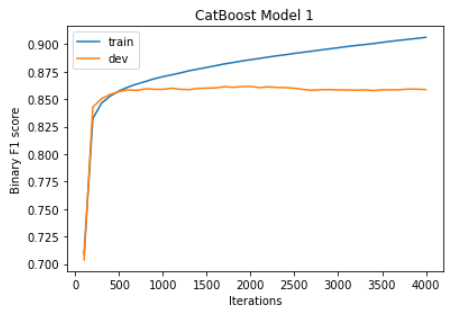}
\caption{Catboost Model 1 train vs dev F1 score}
\end{figure}
\begin{figure}[htbp]
\centering
\includegraphics[width=.9\linewidth]{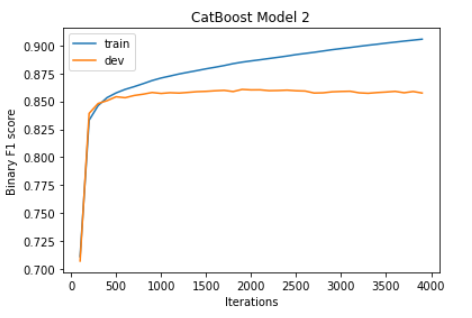}
\caption{Catboost Model 2 train vs dev F1 score}
\end{figure}
\begin{figure}[htbp]
\centering
\includegraphics[width=.9\linewidth]{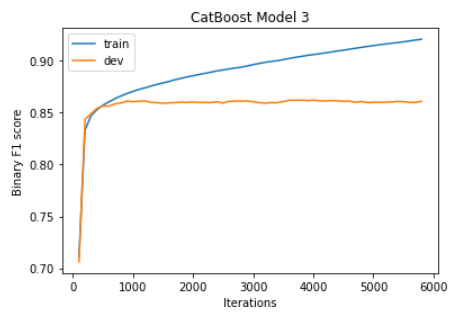}
\caption{Catboost Model 3 train vs dev F1 score}
\end{figure}
\section{\(\mu\)Boost}
\label{sec:org6e99ab9}
In this section we propose a novel effective method to solve text classification
for Indic languages: \textbf{\(\mu\)Boost}. The learnings for its effectiveness are based on
our experiments that we performed in this competition.

As next step to improve
the performance of the model, we tried ensembling the predictions of MURIL and
ensemble CatBoost - µBoost. The motivation behind doing this came from the fact
that there were some languages such as Bhojpuri, Haryanvi and Rajasthani that
were present in the dataset and for which language models like MURIL doesn’t
have adequate representation. Predictions for such languages can be improved
from the trained BoW ensemble CatBoost models. Also, ensembling methods can
sometimes boost the model performance \cite{opitz1999} for
variety of classification tasks.  Here we averaged the individual predicted
probabilities from each of the 3 trained CatBoost models with that of MURIL
model to get the final abuse comment prediction probability. With a probability
threshold of 0.5, the train F1 score shot up to 92.831 and test F1 score to
89.282, which was a considerable improvement in test-set performance.
\section{Threshold tuning}
\label{sec:org78d2295}
The class distribution of abusive to non-abusive comments in dataset was 1:1.12,
which puts it in nearly balanced-class category dataset where there is almost
equal representation from both the classes of labels. We tried tuning the
probability threshold to improve the model’s F1 score by varying it over the
range [0.45,0.55]. The choice of range was based on the fact that the dataset
was nearly balanced in nature. We observed the maximum F1-score at 0.49
probability threshold that gave an F1-score of 89.286.
\begin{figure}[htbp]
\centering
\includegraphics[width=.9\linewidth]{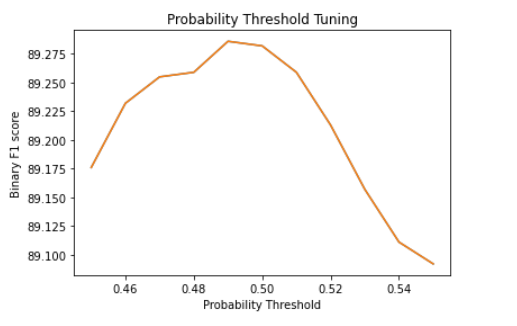}
\caption{F1 score vs classification threshold}
\end{figure}
\section{Future work}
\label{sec:orgab38de7}
The µBoost method significantly improved the model’s performance on unseen data. However, there are many more ways by which it can be improved even further. Since µBoost is made up CatBoost models and MURIL, each of the two can be improved individually to improve the overall performance of the ensemble. Some of the ways could be as follows:
\begin{itemize}
\item Tuning plethora of hyperparameters such as \emph{depth}, \emph{learning\_rate}, \emph{l2\_leaf\_reg} etc. of CatBoost model and \emph{epoch}, \emph{max\_len} etc. of MURIL model via some hyperparameter search method.
\item Increasing the ensemble size of CatBoost models.
\item Using a model-based approach, apart from averaging out, to combine the individual probabilities of the models in the ensemble to improve the performance.
\item Explore other models in the ensemble which can ameliorate the overall model performance.
\end{itemize}

\section{Conclusion}
\label{sec:org278190a}
Table \ref{perf} shows the performance of all our experiments on train and test datasets.
As the result of these set of experiments, we would like to draw the following conclusions:
\begin{itemize}
\item CatBoost model applied on bag of words performs almost at-par with MuRIL. Therefore, we show that CatBoost suitable for  NLP tasks.
\item Ensemble of models, Catboost in this instance, outperform single model based approaches.
\item \(\mu\)Boost outperforms all other component models. Our hypothesis is that \(\mu\)Boost picks up the slack in inputs of languages that are not handled by MuRIL.
\end{itemize}

\bibliographystyle{splncs04}
\bibliography{refs}
\end{document}